\documentclass[10pt,twocolumn,letterpaper]{article}

\usepackage[pagenumbers]{cvpr} 

\usepackage{algorithm}
\usepackage{algorithmic}
\usepackage{tabularx}
\usepackage{multirow}
\usepackage{makecell}
\usepackage{booktabs}
\usepackage{amssymb}
\usepackage{amsmath}

\usepackage[pagebackref,breaklinks,colorlinks]{hyperref}

\def\paperID{5608} 
\def\confName{CVPR}
\def\confYear{2025}

\title{CLDA-YOLO: Visual Contrastive Learning Based \\ Domain Adaptive YOLO Detector}

\author{Tianheng Qiu$^{1,2}$ \;
Ka Lung Law$^{3}$ \;
Guanghua Pan$^{3}$ \;
Jufei Wang$^{1,2}$ \;
Xin Gao$^{4}$ \;
Xuan Huang$^{\ddag}$$^{2}$ \; 
Hu Wei$^{\ddag}$$^{2}$ \; \vspace{+0.1mm}\\
\small $^{\ddag}$ Coresponding Author \vspace{2mm}\\
$^{1}$University of Science and Technology of China\\
$^{2}$Hefei Institutes of Physical Science, Chinese Academy of Sciences\\
$^{3}$SenseTime Research \quad\; $^{4}$Tsinghua University \\
\vspace{-6mm}
\\
}

\begin{document}
\maketitle
\begin{abstract}
    Unsupervised domain adaptive (UDA) algorithms can markedly enhance the performance of object detectors under conditions of domain shifts, thereby reducing the necessity for extensive labeling and retraining. Current domain adaptive object detection algorithms primarily cater to two-stage detectors, which tend to offer minimal improvements when directly applied to single-stage detectors such as YOLO.
    Intending to benefit the YOLO detector from UDA, we build a comprehensive domain adaptive architecture using a teacher-student cooperative system for the YOLO detector. In this process, we propose uncertainty learning to cope with pseudo-labeling generated by the teacher model with extreme uncertainty and leverage dynamic data augmentation to asymptotically adapt the teacher-student system to the environment.
    To address the inability of single-stage object detectors to align at multiple stages, we utilize a unified visual contrastive learning paradigm that aligns instance at backbone and head respectively, which steadily improves the robustness of the detectors in cross-domain tasks.
    In summary, we present an unsupervised domain adaptive YOLO detector based on visual contrastive learning (CLDA-YOLO), which achieves highly competitive results across multiple domain adaptive datasets without any reduction in inference speed.
\end{abstract} 
\begin{figure}[htp]
    \setlength{\abovecaptionskip}{+0.15cm}
    \setlength{\belowcaptionskip}{-0.4cm}
    \centering
    \includegraphics[width=1.0\linewidth]{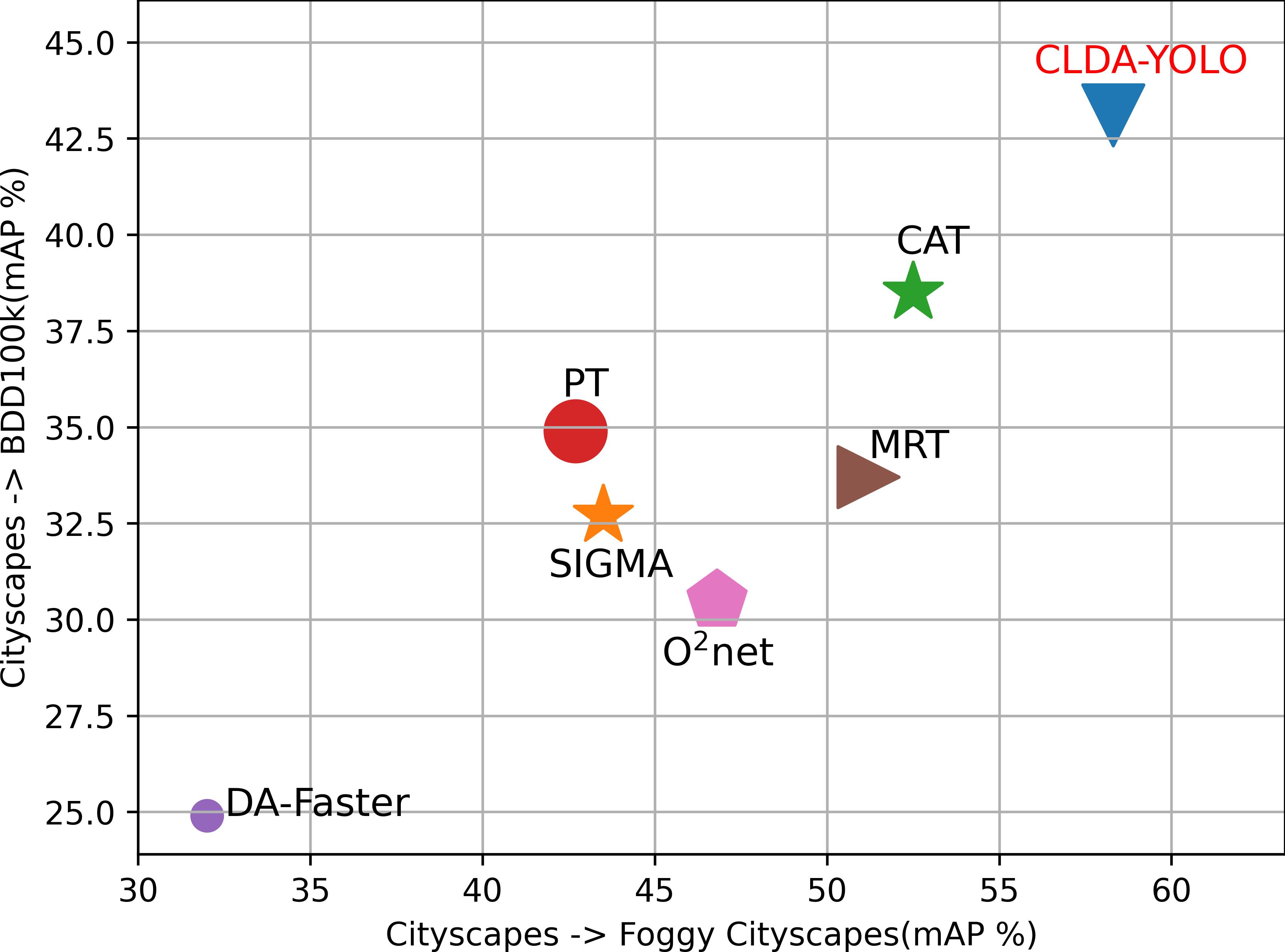}
    \caption{Comparison of mAP@.5 results under two experiment settings, our proposed method obtains the best performance.}
    \label{fig:metrics}
\end{figure}
\section{Introduction}

Object detection is widely used in the real world, and has demonstrated its importance in the fields such as automated driving, industry, and healthcare.

However, before deployment for use, it is often necessary to go through supervised training with accurate and large amounts of manual labeling.
Caused by domain shift~\cite{chen2018domain}, object detection tend to fail when the actual application scenarios are vastly different from the training domain, e.g., in foggy weather, conventional detection models tend to suffer from a large number of misdetections.

To address this problem, some works try to cascade one image restoration or migration model to transform the target domain without labeled data into the source domain first~\cite{liu2022image,dong2020multi}, which not only slows down the inference speed dramatically, but sometimes fails to improve the performance since the images produced by the restoration model do not have a enough confidence~\cite{sun2022rethinking}.
Some other solutions are to train the restoration jointly with the object detection~\cite{wang2022togethernet,huang2020dsnet}, by sharing certain layers of the network to force the detection model to learn implicit information of the target domain. 
However, image restoration and high-level tasks focus on the low-level features and the deeper abstract features respectively, the performance improvement brought by joint training is limited.
A more widely researched approach is known as domain adaptive object detection~\cite{chen2018domain,li2022cross,yu2022mttrans,yoo2022unsupervised}, which makes use of image-to-image translation~\cite{hsu2020progressive, chen2020harmonizing}, domain alignment~\cite{sun2016deep, chen2020homm}, adversarial training~\cite{ganin2016domain}, and other domain adaptive related solutions to greatly improve the performance of detectors at domain shifts without increasing the inference complexity.

In the last few years, we noticed that the researches on unsupervised domain adaptive object detection focus on Faster R-CNN based method~\cite{deng2021unbiased, chen2022learning, kennerley2024cat} or DETR based method~\cite{yu2022mttrans,huang2022aqt,zhao2023masked}.
However, in real-world applications, single-stage object detectors (esp. YOLO~\cite{redmon2016you}) are more widely used, so domain adaptive algorithms for single-stage object detectors are of great research value. 

In this work, we first construct an unsupervised domain adaptive architecture for YOLO, which improves the performance of the detector under domain shift through sufficient cooperative learning between teacher and student model. 
In this process, the pseudo-labels from the teacher model possess strong uncertainty, and the performance of the teacher-student system is limited by the quality of the pseudo-labels~\cite{liu2021unbiased,chen2022label}. To take advantage of this uncertainty and avoid its shortcomings, we design uncertainty learning based on pseudo-labels hierarchy. 
For positive samples, we employ the standard detection loss~\cite{li2020generalized}. For uncertain samples, we opt to utilize the classification confidence of the pseudo-labels as soft labels. As for negative samples, our focus is solely on learning their bounding box loss, ensuring that the False Negative detection boxes retain the capacity to be converted into True Positives during the subsequent cooperative process.
Since teacher-student cooperation is a gradual process, we design adaptive dynamic data augmentation, which gradually varies the strength of the data according to the current degree of stabilization of the teacher-student system and its adaptive ability to the environment.

Compared to domain adaptation for classification tasks, object detection is concerned with aligning more detailed information such as morphology and scale of objects, in addition to the overall domain shift of the image. 
To address this problem, two-stage methods~\cite{chen2018domain} usually perform multi-level alignment at both RPN and ROI, but the inherent paradigm of YOLO reduces object detection to a regression problem, which directly predicts the box information, losing the ability of such multi-level alignment, and is prone to overly relying on the truth labels of the source domains during teacher-student confrontation, resulting in extremely limited performance improvement~\cite{zhang2021domain,hnewa2023integrated}.
From another perspective, existing domain adaptive networks using domain discriminators only for adversarial learning lack theoretical guarantee that two different domain distributions are identical~\cite{long2018conditional, arora2017generalization}, even if the discriminators completely obfuscate the different domains.

For these reasons, we propose to additionally utilize a unified visual contrastive learning paradigm to enforce alignment of instances in target and source domains.
By designing a sigmoid based contrastive loss and domain relevant queues, we align instance features both in backbone and head stage after ROI-Pooling~\cite{ren2015faster}.
Specifically, our optimization goal is to have as much similarity as possible for detection boxes of the same category, regardless of whether they belong to the same domain or not, and conversely, for detection boxes of different categories, we would like them to be as far apart as possible, even if they are under the same domain, which provides the YOLO detector with multiple stages of alignment and enabling performance improvements comparable to other categories of detectors.

Our contributions are set out below:
\begin{itemize}
    \item We build a comprehensive domain adaptive framework for the YOLO detector, capable of aligning student models at multiple stages to achieve stable domain adaptation.
    \item We propose two effective strategies to improve the performance of teacher-student systems.
    We first propose the uncertainty learning strategy, which grades the pseudo-labeling and constructs different losses at different levels to guide the update of system.
    Then we estimate the dynamic differences in the teacher-student system for dynamic data augmentation, gradually increasing the strength of data augmentation as its stability increases.
    \item We propose a multi-stage alignment based on contrastive learning, which utilizes visual contrastive learning to align both the backbone and the head of algorithm, preventing performance degradation caused by over-reliance on source domain labels in the student model.
    \item Without increasing the baseline model complexity, we demonstrate the effectiveness of the proposed algorithm in multiple dataset settings.
\end{itemize}

\begin{figure*}[ht]
    \setlength{\abovecaptionskip}{+0.6mm}
    \setlength{\belowcaptionskip}{-0.3cm}
    \centering
    \includegraphics[width=1.\linewidth]{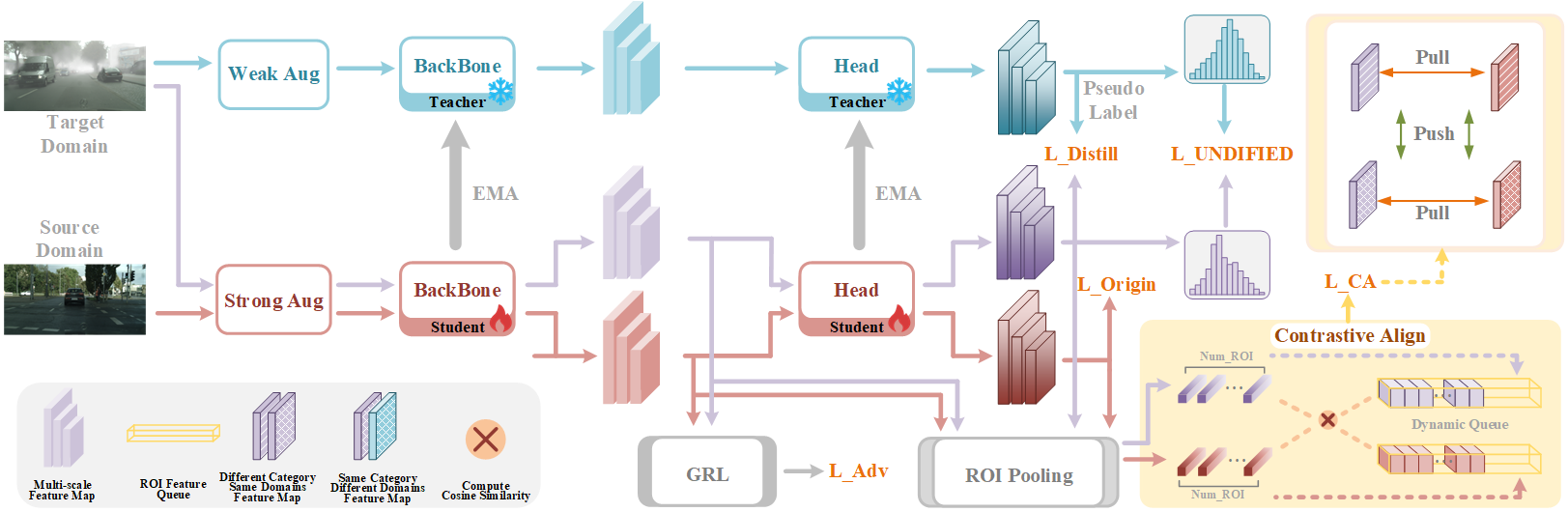}
    \caption{The overall architecture of the proposed CLDA-YOLO. Following the teacher-student cooperative learning architecture, we built a domain adaptive architecture for YOLO detector, where the teacher model generates pseudo-labels to compute distillation loss and uncertainty loss, the student model computes the source-domain supervised loss and distillation loss in addition to the contrastive alignment loss that we additionally set in order to enable the detector to perform a coherent alignment on the backbone and the head, respectively.}
    \label{fig:overall}
\end{figure*}
\section{Related Works}

\noindent\textbf{Domain adaptive object detection}. Domain Adaptive Object Detection (DAOD) is used to solve the domain shift in object detection. \cite{chen2018domain} first proposed to additionally train a domain discriminator in Faster R-CNN for adversarial feature alignment, and~\cite{hu2023dagl,deng2021unbiased,munir2021ssal,li2022cross} followed this paradigm to carry out extensive research, and since then the self-training technique of mutual learning between teacher and student used in~\cite{li2022cross, chen2022learning}, the image-to-image translation technique in~\cite{deng2021unbiased, chen2020harmonizing, hsu2020progressive} have all been extensively researched. 

Numerous DETR-basaed DAOD methods have also emerged after the rise of DETR~\cite{zhu2020deformable}, among the more advanced works~\cite{huang2022aqt, yu2022mttrans, zhao2023masked, gong2022improving},~\cite{huang2022aqt} proposed the use of adversarial tokens to improve the performance of transformer-based detectors, and~\cite{zhao2023masked} introduced Masked Auto-Encoder into domain adaptation, and utilized techniques such as retraining to achieve competitive results. 

The use of single-stage object detectors in DAOD, represented by the FCOS detector~\cite{tian2019fcos}, has also received increasing attention, and~\cite{li2022sigma} introduced Graph Network into DAOD to achieve alignment between different domains through Graph Matching.~\cite{deng2023mharonious} proposed a harmony-based pseudo-label weighting method to fully utilize the prediction results of teacher networks. 
However, another popular single-stage object detector YOLO, has not been sufficiently emphasized in this task.~\cite{zhang2021domain} introduced instance-level alignment into the YOLO detector to follow the paradigm proposed by~\cite{chen2018domain}, and~\cite{hnewa2023integrated} investigated a stronger domain discriminator based on the multi-scale behavior of the YOLO detector, but the performance gains from the approach they adopted are limited. 
In this study, we propose a domain-adaptive architecture for YOLO that enhances its competitiveness with other DAOD methods.

\noindent\textbf{Visual Contrastive Learning}. 
Contrastive learning enables models to implicitly learn the similarities and differences between samples, and many excellent contrastive learning algorithms have been proposed based on instance discrimination tasks~\cite{wu2018unsupervised,chen2020simple,chen2020improved}. 
To boost the number of negative samples,~\cite{he2020momentum} generalizes the contrastive learning method to a dictionary query problem, proposed to maintain a dynamic queue to increase the number of negative samples, and guarantees the consistency of negative samples by a momentum encoder. 
Extending self-supervised contrastive learning to supervised,~\cite{khosla2020supervised} proposed category-based contrastive loss, which provides theoretical guarantees for extending contrastive learning to more visual downstream tasks. 
In this work, we utilize contrastive learning for alignment between different domains, and unlike numerous visual contrastive learning efforts, since the positive and negative samples are extremely unbalanced in object detection, the need to compute loss for multiple positive and negative samples simultaneously shifts it from a multi-category to a multi-label task, so we use sigmoid-based contrastive learning loss instead of the InfoNCE~\cite{he2020momentum,khosla2020supervised}.

\section{Proposed Method}
In order to align our proposed domain adaptive algorithm to the most advanced YOLO object detection algorithm, we set the detection head as an anchor-free detector to ensure that the proposed algorithm can be adapted by simple changes. 
Specifically, we use the positive and negative sample allocation strategy~\cite{feng2021tood} to match the detection boxes, and utilize the DFL~\cite{li2020generalized} to model the uncertainty of the detection boxes and unify the confidence level of training and inference, which is the same as the state-of-the-art anchor-free single-stage object detection algorithms\cite{li2023yolov6, yolov8_ultralytics, wang2024yolov9}. 
The overall architecture of the algorithm is shown in Fig.~\ref{fig:overall}.

\subsection{Baseline Architecture}
Follow~\cite{li2022cross, liu2021unbiased}, we input the strongly augmented source and target domain images into the student model respectively, and the weakly augmented target domain image into the teacher model to generate pseudo-labels, then the teacher model parameters $\theta_t$ integrates the parameters of the student model $\theta_s$ by means of an exponential moving average (EMA) as follows:
\begin{equation}
    \theta_t \leftarrow \alpha \theta_t + (1-\alpha)\theta_s
\end{equation}
where $\alpha$ is the decay controls the update momentum.

Given the labeled source domain dataset $D_s=\{X_s;Y_s\}$, and an unlabeled target domain dataset $D_t=\{X_t\}$, where $X_s$ and $X_t$ represent the images of the corresponding domains, $Y_s=\{B_s;C_s\}$ denotes the label of the corresponding image, which contains coordinate boxes supervised $B_s$ and categories supervised $C_s$.
For labeled source-domain samples, student model first compute the object detection loss $\mathcal{L}_{sup}$ using the conventional:
\begin{equation}
    \mathcal{L}_{sup}(X_s, Y_s) = \mathcal{L}_{cls}(X_s, C_s)+\mathcal{L}_{dfl}(X_s, B_s)+\mathcal{L}_{iou}(X_s, B_s)
\end{equation}
where $\mathcal{L}_{cls}$ denotes the classification loss of the detection boxes, the border probability loss $\mathcal{L}_{dfl}$ and the IoU loss $\mathcal{L}_{iou}$ together constitute the regression loss of the detection boxes (same below).

After that, we will perform uncertainty learning based on the pseudo labels $\hat{Y}_t=\{\hat{B}_t;\hat{C}_t\}$ produced by the teacher model to compute $\mathcal{L}_{distill}$.

\noindent\textbf{Uncertainty learning}. For sample of the target domain that requires pseudo-labeling supervision, we use uncertainty-based detection loss. 
Pseudo-label assignment strategies have been extensively studied in semi-supervised object detection~\cite{xu2021end, tang2021humble, chen2022label, liu2021unbiased, xu2023efficient}, and in order to take advantage of the uncertainty introduced by pseudo-labels, we leverage the simplest label assignment strategy, which divides the detection frame into three categories by setting hyper-parameter thresholds $p_l$ and $p_h$.
Specifically, labels larger than the preset confidence $p_h$ are labeled as positive samples, those with confidence in the interval $[p_l, p_h]$ are labeled as uncertain samples, and otherwise, they are noted as negative samples.

Since non-maximum suppression~\cite{neubeck2006efficient} during inference is now related to classification confidence~\cite{li2020generalized}, we set the classification probability of positive samples to 1.0 to amplify their effect as pseudo-labels. 
The classification scores of the uncertainty samples are directly replaced with the original labels as a way of suppressing possible false samples in the student model, which can be expressed as:
\begin{equation}
    \begin{aligned}
        \mathcal{L}_{cls}^{'}(X_{t},\hat{C}_{t}) = & \sum_{i,j} \mathbb{I}_{[p_l<\hat{P}_{t}^{ij}<p_h]}BCE(X_t^{ij}, \hat{P}_{t}^{ij}) \\
        &+ \mathbb{I}_{[\hat{P}_{t}^{ij}\geq p_h]}BCE(X_t^{ij},1)
     \end{aligned}
\end{equation}
where $\hat{P}_t^{ij}$ is the confidence of the teacher model to generate the corresponding location detection box, before which $X_t^{ij}$ and $\hat{Y}_t^{ij}$ have been matched by the label allocator, $\mathbb{I}$ denotes the indicator function, which outputs a 1 when the condition is satisfied, and vice versa.

At the same time, not changing the regression loss of the detection boxes ensures that existing False Negative detection boxes have a chance to be converted to True Positive in the future.
In addition, the module retains a certain amount of migration performance since the classification loss is used in almost all object detection algorithms.

For detection boxes with negative sample labeling added, the classification loss that would compute them as pseudo-labels was no longer considered due to their extremely low confidence. Since these detection boxes have low confidence for both classification and border regression, the same hard loss as for the other types cannot be used to regress the borders, and we use JS-Divergence to measure the distance of the distribution of these pseudo-labels with very low confidence from the output of the student model, which can be expressed as:
\begin{equation}
    \begin{aligned}
        \mathcal{L}_{ng}(X_t, \hat{B}_{t}) &= \sum_{ij}\mathbb{I}_{[\hat{P}_{t}^{ij}<p_l]}JS(X_{t}^{ij},\hat{B}_{t}^{ij})
     \end{aligned}
\end{equation}
where $JS$ represents the computation of JS-Divergence, so that the final pseudo-label distillation loss is:
\begin{equation}
    \begin{aligned}
    \mathcal{L}_{distill}(X_t, \hat{Y}_t)&=\mathcal{L}_{cls}^{'}(X_t, \hat{C}_t)+\mathcal{L}_{dfl}(X_t, \hat{B}_t) \\
    &+\mathcal{L}_{iou}(X_t, \hat{B}_t)+\mathcal{L}_{ng}(X_t, \hat{B}_{t})
    \end{aligned}
\end{equation}

\noindent\textbf{Dynamic Data Augmentation}. In the early stage of training, the model is poorly adapted to the target domain, we propose a dynamic data augmentation strategy to make the teacher-student model dynamically adapt to the environment. As the model's adaptability to the target domain gradually increases, the intensity of data augmentation is gradually increased. Our proposed dynamic parameter is composed of the JS-Divergence of the teacher and student models with respect to the distribution of detection boxes in the target domain and the entropy weighting of the pseudo-labeling of the teacher model, as shown in Eq.6.
\begin{equation}
    \begin{aligned}
        P_{t}^{k} = &\left(\frac{1}{hw}\sum_{ij}JS(reg^{stu}_{ij},reg^{tea}_{ij})\right)_{t}^{\gamma} \\
        P_{t}^{gain} &= \alpha P_{t-1}^{gain} + (1-\alpha)\frac{\bar P_{t_0}}{\bar P_{t}}
    \end{aligned}
\end{equation}
where $t$ denotes the number of update count, $\bar P_{t}$ is the average value of all batches at this update $t$, $\gamma$ is control factors, and $reg$ represents the model prediction at each anchor $i,j$.
We utilize JS-Divergence again for measuring the adaptability of the teacher-student system to the current environment, when the divergence decreases we consider that it becomes more adaptable to the current environment.
\subsection{Contrastive Learning-based Alignment}
\begin{figure}[ht]
    \setlength{\abovecaptionskip}{-0.02mm}
    \setlength{\belowcaptionskip}{-0.5cm}
    \centering
    \includegraphics[width=1.\linewidth]{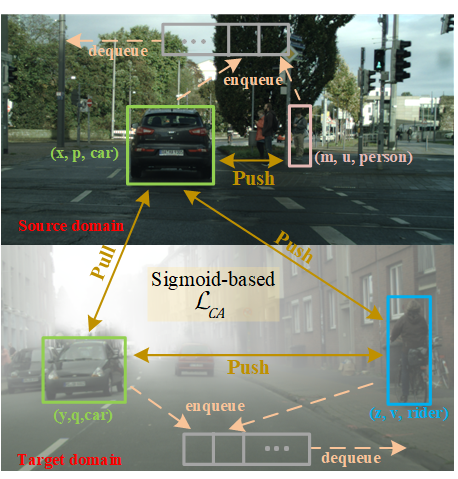}
    \caption{Simple schema of our Contrastive Learning-based Alignment. For each box, we provide tuple to describe it, which means (features, confidence, category). The queue update is executed after the whole batch has been computed.}
    \label{fig:cons}
\end{figure}
To enable the performance of multi-stage alignment in the YOLO detector, we propose Visual Contrastive Learning-based Alignment (CA) method that forces instances in target and source domains to be aligned with conditional probability distributions, as shown in Fig.~\ref{fig:cons}.

For feature alignment, we retain the same feature-level Gradient Reverse Layer (GRL)~\cite{ganin2015unsupervised} as the advanced domain adaptive object detection algorithms~\cite{chen2018domain,li2022cross}, which is used to align edge distributions between the different domains. The domain adversarial loss $\mathcal{L}_{adv}$ is represented as follows:
\begin{equation}
    \begin{aligned}
        \mathcal{L}_{adv} &= \min_{E}\max_{D} \mathcal{L}_{dis} \\
        \mathcal{L}_{dis} &= -d \log D(E(X)) - (1 - d) \log(1 - D(E(X)))
    \end{aligned}
\end{equation}
wherein the domain discriminator $D$ determines the domain label of features generated by the encoder $E$.

Since domain alignment is global and limited by the number of samples generated within a single batch, direct comparison learning within a batch cannot compare to the global objective, but instead tends to make comparison learning fall into a local optimum, so we maintain a dynamic circular queue to store the detection boxes generated in multiple batch~\cite{he2020momentum}, and when enough samples are stored in the queue, comparison learning will learn global and uniform conditional probability information. Unlike the queue appearing in~\cite{he2020momentum}, we do not maintain a Momentum Updated Encoder additionally, partly because in this task, positive and negative samples are not absolute in the batch, and partly because object detection is an intensive output task , which means the queue is updated very fast, with high consistency of samples.

Specifically, in the m-th layer alignment of the backbone guided by the m-th detector (same for the head), we maintain queues $Q_s^m\in (K,l^m)$ and $Q_t^m\in (K,l^m)$ of length $K$ for the source and target domains, respectively.
For the number of $n_1$ prediction box in the source domain and the $n_2$ in the target domain, we project them to the corresponding layers of the backbone by ROI-Pooling~\cite{ren2015faster} projected onto the feature maps $f_{s}^m$ and $f_{t}^m$ of the corresponding layer of the backbone to get instance feature $x_s^m \in (n_1,l^m)$, $x_t^m \in (n_2,l^m)$, where $l^m$ denotes the tuple shown in Fig.~\ref{fig:cons}.
Then computed with the queue of the cross domain and the queue of the home domain to get the similarity matrix respectively, and based on the matrix to compute the contrastive learning loss bootstrap for cross-domain alignment, and finally we put the $x_s^m$ into $Q_s^m$ and $x_t^m$ into $Q_t^m$ to complete the update of the queue.

In the setting of the object detection task, visual contrastive learning will no longer be an instance discrimination task, and the use of softmax-based InfoNCE for a given sample $x_i$ for which there are multiple positive samples corresponding to it is underpowered, as it suppresses other similar or similar categories. 
In simple terms, we need to transform a multi-category learning task into a multi-label learning task.
Inspired by \cite{zhai2023sigmoid}, we design Sigmoid Contract Align Loss for the object detection task, which independently calculates the cosine similarity of each sample pair and computes the loss, as represented by the following loss function:
\begin{equation}
    \begin{aligned}
        \mathcal{L}_{CA} = -\frac{1}{K}\sum_{i=1}^{K}\sum_{j=1}^{n}(1-p_i^{\frac{\alpha}{2}}q_j^{\frac{\beta}{2}})\log{\frac{1}{1+e^{-x_iy_j \times mask_{ij} \times \tau}}}
    \end{aligned}
\end{equation}
where $mask_{ij}$ is the mask of the equal category of sample $i$ and $j$, we set the position to 1 for the same category and -1 for different categories.
Temperature factor $\tau = lnT$, where we use a log function to prevent the problem of gradient vanishing caused by the rapid rise in temperature at the beginning of training. 
$p_i$,$q_j$ denote the classification confidence of detection box $i$ in the queue and detection box $j$ in the batch, respectively.
A succinct and clear pseudo-code for $\mathcal{L}_{CA}$ can be found in the Appendix.
\begin{table*}[h]
    \setlength{\abovecaptionskip}{+0.1cm}
    \setlength{\belowcaptionskip}{-0.31cm}
    \setlength{\tabcolsep}{4pt}
    \centering
    \begin{tabular*}{0.86\textwidth}{l||c||cccccccc||c}
    \toprule
    Method & Detector & person  & rider & car  & truck & bus & train & mcycle & bicycle & mAP@.5 \\
    \midrule
    Faster-RCNN~\cite{ren2015faster} & Faster & 26.9 & 38.2 & 35.6 & 18.3 & 32.4 & 9.6 & 25.8 & 28.6 & 26.9 \\
    PT~\cite{chen2022learning}  & Faster  & 40.2 & 48.8 & 59.7 & 30.7 & 51.8 & 30.6 & 35.4 & 44.5 & 42.7 \\
    AT~\cite{li2022cross} & Faster    & 45.3 & 55.7 & 63.6 & 36.8 & 64.9 & 34.9 & 42.1 & 51.3 & 49.3 \\
    CMT~\cite{cao2023contrastive}  & Faster & 45.9 & 55.7 & 63.7 & 39.6 & 66.0 & 38.8 & 41.4 & 51.2 & 50.3 \\
    MILA~\cite{krishna2023mila} & Faster & 45.6 & 52.8 & 64.8 & 34.7 & 61.4 & 54.1 & 39.7 & 51.5 & 50.6 \\
    CAT~\cite{kennerley2024cat} & Faster  & 44.6 & 57.1 & 63.7 & 40.8 & 66.0 & 49.7 & 44.9 & 53.0 & 52.5\\
    \midrule
    Def-DETR~\cite{zhu2020deformable} & Def-DETR & 37.7 & 39.1 & 44.2 & 17.2 & 26.8 & 5.8 & 21.6 & 35.5 & 28.5 \\
    MTTrans~\cite{yu2022mttrans} & Def-DETR & 47.7 & 49.9 & 65.2 & 25.8 & 45.9 & 33.8 & 32.6 & 46.5 & 43.4 \\
    AQT~\cite{huang2022aqt} & Def-DETR  & 49.3 & 52.3 & 64.4 & 27.7 & 53.7 & 46.5 & 36.0 & 46.4 & 47.1 \\
    O$^2$net~\cite{gong2022improving} & Def-DETR  & 48.7 & 51.5 & 63.6 & 31.1 & 47.6 & 47.8 & 38.0 & 45.9 & 46.8 \\
    MRT~\cite{zhao2023masked} & Def-DETR  & 52.8 & 51.7 & 68.7 & 35.9 & 58.1 & 54.5 & 41.0 & 47.1 & 51.2 \\
    \midrule
    FCOS~\cite{tian2019fcos} & FCOS & 36.9 & 36.3 & 44.1 & 18.6 & 29.3 & 8.4 & 20.3 & 31.9 & 28.2 \\
    SSAL~\cite{munir2021ssal} & FCOS & 45.1 & 47.4 & 59.4 & 24.5 & 50.0 & 25.7 & 26.0 & 38.7 & 39.6 \\
    SIGMA~\cite{li2022sigma} & FCOS & 46.9 & 48.4 & 63.7 & 27.1 & 50.7 & 35.9 & 34.7 & 41.4 & 43.5 \\
    HT~\cite{deng2023mharonious} & FCOS   & 52.1 & 55.8 & 67.5 & 32.7 & 55.9 & 49.1 & 40.1 & 50.3 & 50.4 \\
    \midrule
    YOLOv5-L~\cite{Jocher_YOLOv5_by_Ultralytics_2020} & YOLOv5 & 46.5 & 49.0 & 51.9 & 24.2 & 37.2 & 8.5 & 30.3 & 39.2 & 35.9 \\
    \citeauthor{zhang2021domain}~\cite{zhang2021domain} & YOLOv3 & 29.5 & 27.7 & 46.1 & 9.1 & 28.2 & 4.5 & 12.7 & 24.8 & 35.1 \\
    MS-DAYOLO~\cite{hnewa2023integrated} & YOLOv4 & 39.6 & 46.5 & 56.5 & 28.9 & 51.0 & 45.9 & 27.5 & 36.0 & 41.5 \\
    SSDA-YOLO~\cite{zhou2023ssda} & YOLOv5     & 60.6 & 62.1 & 74.3 & 37.8 & 63.0 & 48.0 & 47.4 & 53.6 & 55.9 \\

    \textbf{CLDA-YOLO}$^*$ & YOLOv5 & 51.7 & 55.1 & 67.7 & 35.0 & 62.7 & \textbf{56.5} & 39.4 & 44.9 & 51.6 \\
    \textbf{CLDA-YOLO}\ & YOLOv5 & \textbf{61.6} & \textbf{64.5} & \textbf{74.1} & \textbf{41.7} & \textbf{66.8} & 48.4 & \textbf{50.9} & \textbf{53.8} & \textbf{57.7} \\
    \bottomrule
    \end{tabular*}
    \caption{Domain Adaptive Object Detection Results in Cityscapes $\rightarrow $ Foggy Cityscapes Setting, $^*$ indicates that we used input images with a resolution of 640, otherwise 960. The average precision (AP, \%) on all classes is presented.}
    \label{tab:1}
\end{table*}
\begin{table*}[h]
    \setlength{\abovecaptionskip}{+0.1cm}
    \setlength{\belowcaptionskip}{-0.4cm}
    \setlength{\tabcolsep}{3.3pt}
    \centering
    \begin{tabular*}{0.8\textwidth}{l||c||ccccccc||c}
    \toprule
    Method & Detector & person & rider & car  & truck & bus & mcycle & bicycle & mAP@.5 \\
    \midrule
    Faster-RCNN~\cite{ren2015faster} & Faster & 28.8 & 25.4 & 44.1 & 17.9 & 16.1 & 13.9 & 22.4 & 24.1 \\
    DA-Faster~\cite{chen2018domain} & Faster & 28.9 & 27.4 & 44.2 & 19.1 & 18.0 & 14.2 & 22.4 & 24.9 \\
    PT~\cite{chen2022learning}   & Faster & 40.5 & 39.9 & 52.7 & 25.8 & 33.8 & 23.0 & 28.8 & 34.9 \\
    CAT~\cite{kennerley2024cat}   & Faster & 44.6 & 41.5 & 61.2 & 31.4 & 34.6 & 24.4 & 31.7 & 38.5 \\
    Def DETR~\cite{zhu2020deformable} & Def-DETR & 38.9 & 26.7 & 55.2 & 15.7 & 19.7 & 10.8 & 16.2 & 26.2 \\
    MTTrans~\cite{yu2022mttrans} & Def-DETR & 44.1 & 30.1 & 61.5 & 25.1 & 26.9 & 17.7 & 23.0 & 32.6 \\
    SIGMA~\cite{li2022sigma} & Def-DETR & 46.9 & 29.6 & 64.1 & 20.2 & 23.6 & 17.9 & 26.3 & 32.7 \\
    MRT~\cite{zhao2023masked}  & Def-DETR & 48.4 & 30.9 & 63.7 & 24.7 & 25.5 & 20.2 & 22.6 & 33.7 \\
    YOLOv5-L~\cite{Jocher_YOLOv5_by_Ultralytics_2020} & YOLOv5 & 50.7 & 34.1 & 66.8 & 24.9 & 25.4 & 24.6 & 28.7 & 31.9  \\
    \textbf{CLDA-YOLO}\,(ours) & YOLOv5 & \textbf{61.4} & \textbf{48.3} & \textbf{75.1} & \textbf{41.0} & \textbf{41.7} & \textbf{39.3} & \textbf{38.9} & \textbf{43.2} \\
    \bottomrule
    \end{tabular*}
    \caption{Domain Adaptive Object Detection Results in Cityscapes $\rightarrow $ BDD100k Setting. Following the setup of our previous work, we ignored the "train" label, the average precision (AP, \%) of the rest categories is presented.}
    \label{tab:2}
\end{table*}
\begin{figure*}[htbp]  
    \setlength{\abovecaptionskip}{+0.01cm}
    \setlength{\belowcaptionskip}{-0.5cm}
    \centering
    \includegraphics[width=1.\linewidth]{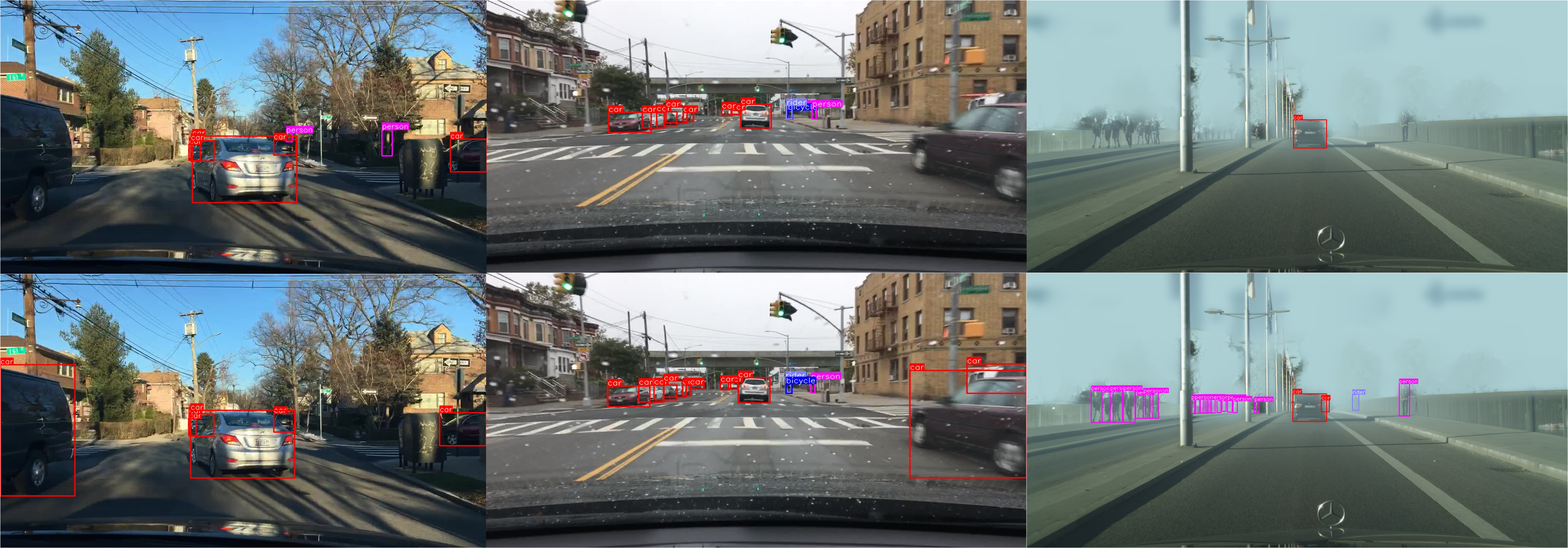}
    \caption{Comparison of CLDA-YOLO's prediction visualization, with images of normal weather, rainy day, and foggy day, from left to right, with Source-Only model's detection results in the first row, and CLDA-YOLO in the second row.}
    \label{fig:ap}
\end{figure*}

In $\mathcal{L}_{CA}$, we set a confidence weight for each sample pair because, given that the detection boxes have difficulty and the distribution of categories is not balanced in the vast majority of cases, in which direct comparative learning would bias the overall loss towards the samples with a large number of appearance, losing the so-called uniformity.
In summary, we obtained the similarity matrices of the regions of interest of backbone and head with the dynamic queue separately on a layer-by-layer basis and calculated their positive and negative sample losses using $\mathcal{L}_{CA}$ so that the single-stage detectors can also align instances in both parts.

\section{Experiments}
\subsection{Datasets and Evaluation Settings}
We used the mean average precision at an IoU threshold of 0.5 (mAP@.5) as the metric for evaluation.
According to~\cite{zhao2023masked, kennerley2024cat, saito2019strong, zhou2023ssda}, we set up the following sets of domain shift object detection experiments.

\noindent\textbf{Cityscapes to Foggy Cityscapes}. 
Cityscapes~\cite{cordts2016cityscapes} is a urban street scenes dataset containing 2975 training images and 500 validation images. Foggy Cityscapes~\cite{sakaridis2018semantic} is a foggy scenes dataset synthesized by Cityscapes. In this setting, we choose Cityscapes as the source domain and Foggy Cityscapes highest density synthetic haze (0.02) as the target domain.

\noindent\textbf{Cityscapes to BDD100k(daytime)}. BDD100k~\cite{yu2020bdd100k} is an automated driving dataset, we use its daytime portion as the target domain, which contains a total of 36,728 training images and 5,258 validation images. For cross-domain training, we align categories of BDD100k toward Cityscapes. 

\noindent\textbf{Sim10k to Cityscapes}. Sim10k~\cite{johnson2016driving} is a synthetic urban street scenes dataset using the game engine of GTA-V. It contains a total of 10,000 images, and we use it as the source domain and Cityscapes as the target domain, with only the detection boxes of the car category.

\noindent\textbf{KITTI to Cityscapes}. KITTI~\cite{geiger2012we} is a real-world autonomous driving dataset containing a total of 7,481 training images, which we use as the source domain and Cityscapes as the target domain, and use only the detection boxes of the car category.
\begin{table}[htbp]
    \setlength{\abovecaptionskip}{+0.01cm}
    \setlength{\belowcaptionskip}{-0.5cm}
    \centering
    \begin{tabular*}{0.35\textwidth}{l||cc}
    \toprule
    \multirow{2}{*}{Method} & \multicolumn{2}{c}{carAP@.5} \\
    & k $\rightarrow $ c & s $\rightarrow $ c \\
    \midrule
    DA-Faster~\cite{chen2018domain} & 41.9 & -\\
    EPM~\cite{hsu2020every} & 45.0 & 51.2\\
    SSAL~\cite{munir2021ssal} & 45.6 & 51.8\\
    PT~\cite{chen2022learning} & \textbf{60.2} & 55.1 \\
    \citeauthor{zhang2021domain}~\cite{zhang2021domain} & 54.0 & 50.9 \\
    MTTrans~\cite{yu2022mttrans} & - & 57.0 \\
    MRT~\cite{zhao2023masked} & - & \underline{62.0} \\
    MS-DAYOLO~\cite{hnewa2023integrated} & 47.6 & - \\
    CLDA-YOLO(ours) & \underline{57.2} & \textbf{66.5} \\
    \midrule
    \end{tabular*}
    \caption{Comparison of experimental results between KITTI $\rightarrow $ Cityscapes and SIM10k $\rightarrow$ Cityscapes, only the average precision (AP, \%) of "car" is reported in the table.}
    \label{tab:3}
\end{table}
 
\subsection{Implementation Details}
We first perform burn-in training for 150 epochs, and then perform teacher-student adversarial training, with a total of 400 epochs for training process. We utilize SGD optimizer with a learning rate of 0.01, and the detailed training hyperparticipation~\cite{yolov8_ultralytics} is kept consistent. We performed all experiments under NVIDIA V100 GPUs with batch-size 64.
\subsection{Comparison with State-of-the-Art Methods}
\noindent\textbf{Weather Adaptation}.
We first validate the performance of the algorithm on the most widely used dataset setup Cityscapes to Foggy Cityscapes in Tab.~\ref{tab:1}. Our proposed CLDA-YOLO improves the mAP by 5.2\% over the previous state-of-the-art model CAT, obtaining very competitive results even for small sizes of the model.
Compared to competitive SSDA-YOLO, our proposed algorithm no longer needs to pre-train a domain translator, but gains higher mAP, representing a wider range of applications.
This shows that CLDA-YOLO has a good adaptability to domain shifts under severe weather conditions.

\noindent\textbf{Scene Adaptation}.
Tab.\ref{tab:2} shows the performance of the domain adaptive algorithm across different scenarios, where CLDA-YOLO beats the previous highest mAP of 38.5 with 43.2 and outperforms the average precision in all categories.

\noindent\textbf{Single-Class Adaptation}.
In the single-domain adaptation with only the automobile category, CLDA-YOLO achieved 66.5 AP in the Sim10k to Cityscapes domain adaptation, which is 4.5\% higher than the previous sota.
\subsection{Ablation Study}
\begin{figure*}[htbp]
    \setlength{\abovecaptionskip}{+0.1cm}
    \setlength{\belowcaptionskip}{-0.25cm}
   \centering
   \begin{minipage}{0.99\textwidth}
      \centering
      \includegraphics[width=0.33\textwidth]{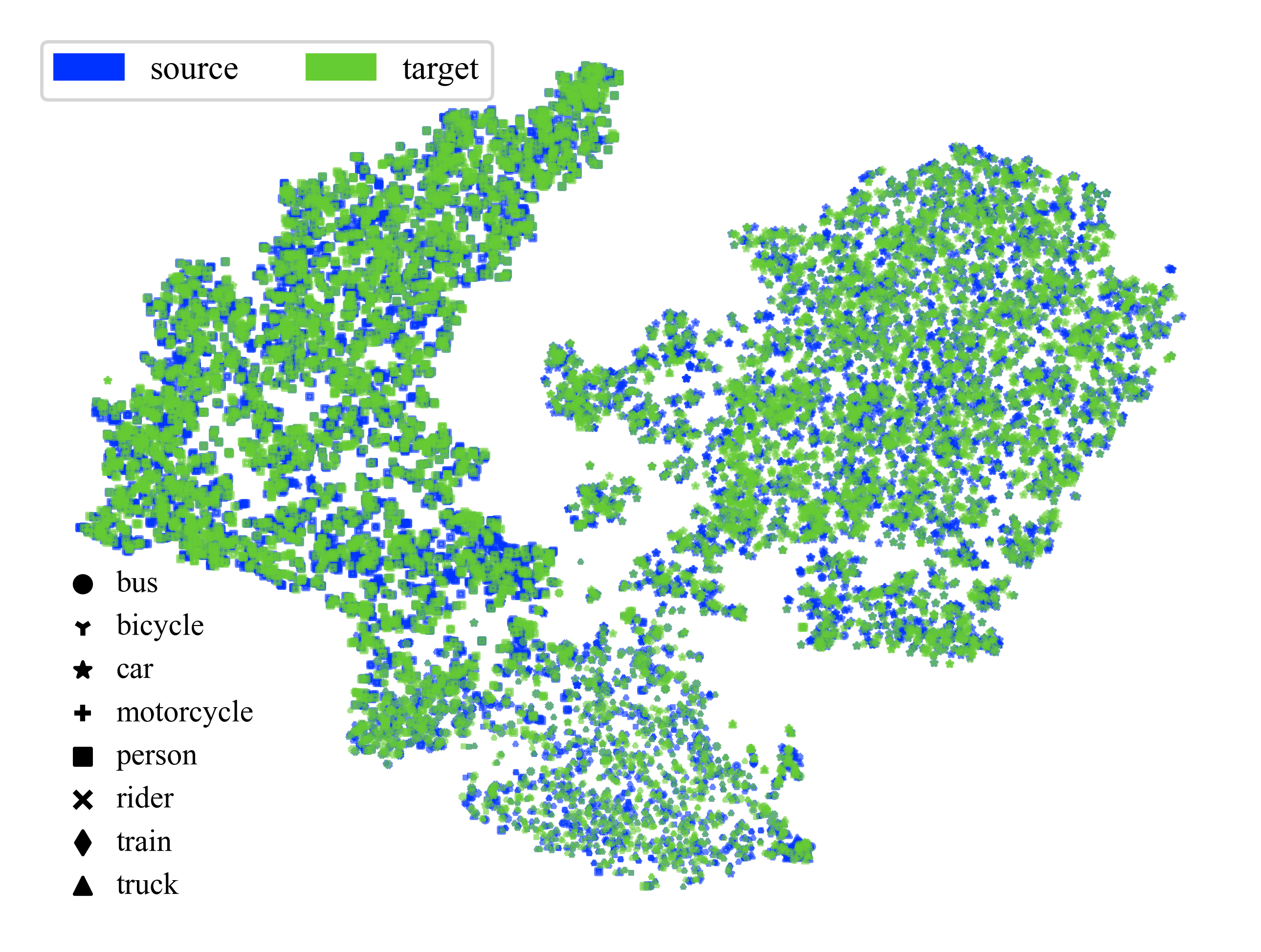}
      \includegraphics[width=0.33\textwidth]{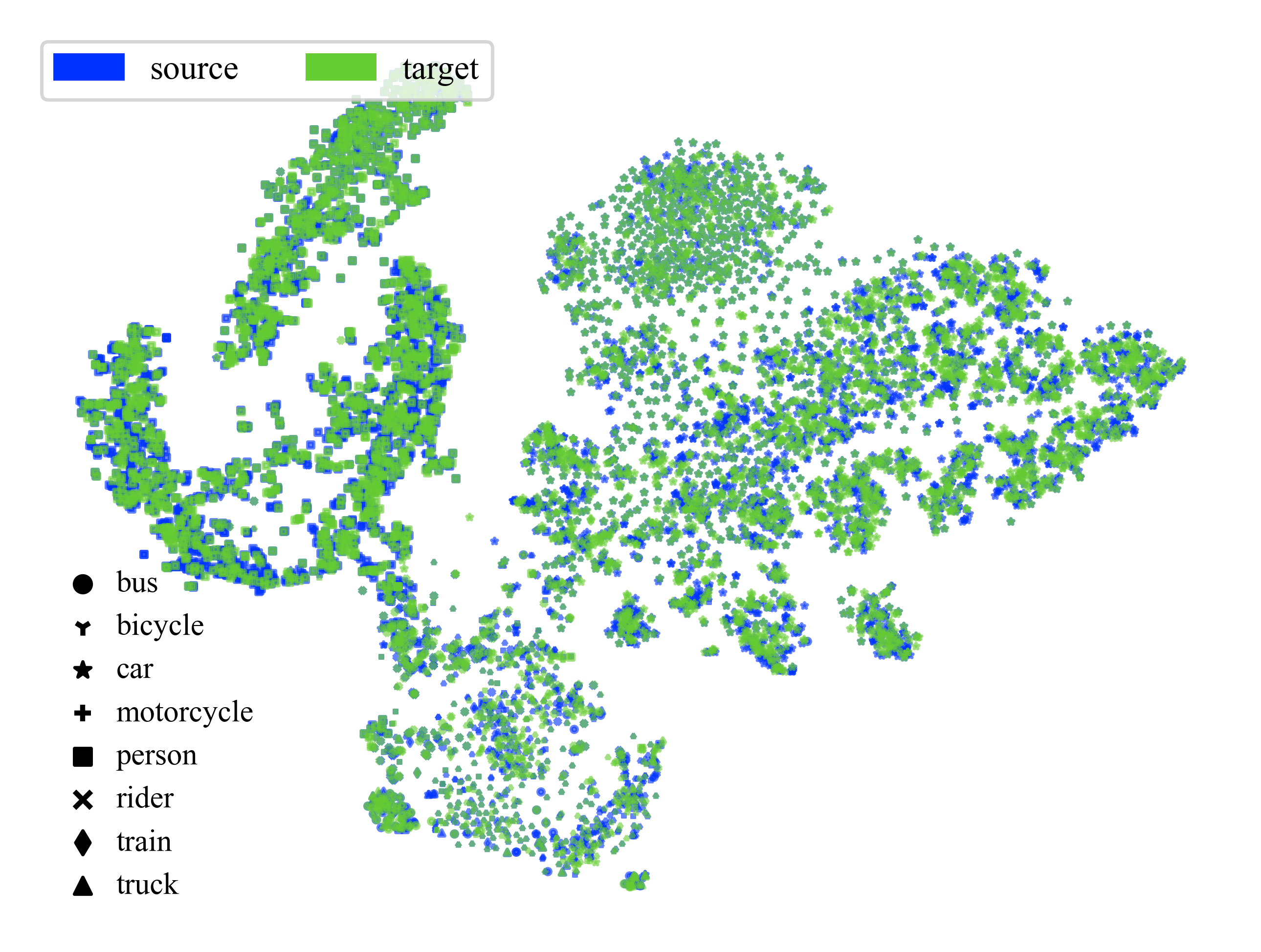}
      \includegraphics[width=0.33\textwidth]{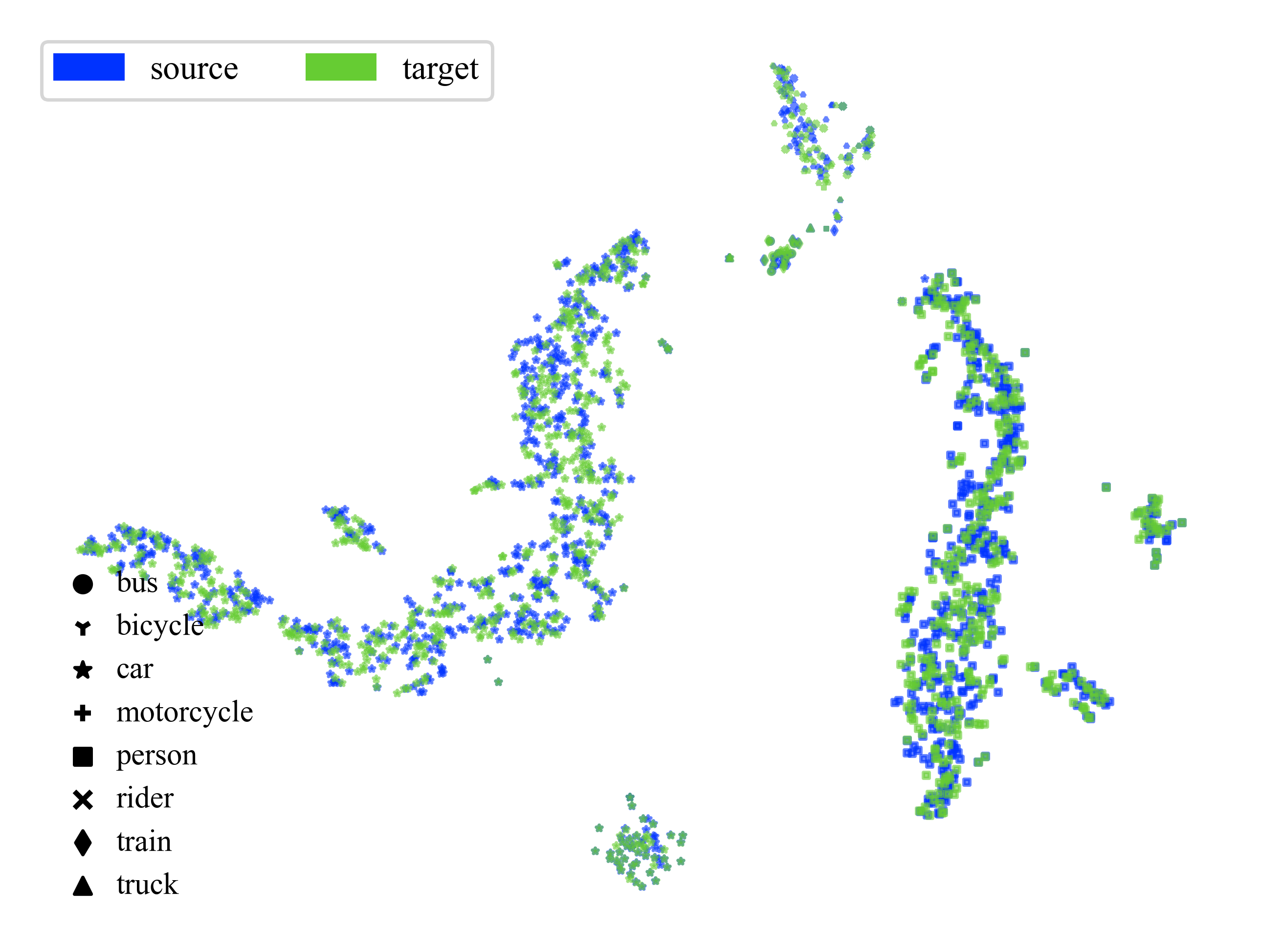}
  \end{minipage}
  \caption{Feature visualization of Cityscapes $\rightarrow $ Foggy Cityscapes by T-SNE, which generated by each detector head. Categories and domains are distinguish by marker and color respectively. Zoom in for more detailed view.}
  \label{fig:tsne}
\end{figure*}
\begin{table*}[htbp]
    \setlength{\abovecaptionskip}{+0.1cm}
    \setlength{\belowcaptionskip}{-0.25cm}
    \centering
    \begin{tabular*}{0.9\textwidth}{ccccccccccc}
    \toprule
    Method & Setting & person  & rider & car  & truck & bus & train & mcycle & bicycle & mAP@.5 \\
    \midrule
    YOLOv8-L      & c $\rightarrow$ f & 55.6 & 60.4 & 67.9 & 33.2 & 50.2 & 19.4 & 34.8 & 51.6 & 46.6 \\
    CLDA-YOLOv8-L & c $\rightarrow$ f & 61.8 & 65.9 & 76.2 & 37.1 & 66.1 & 55.8 & 50.3 & 55.8 & 58.6 \\
    \midrule
    YOLOv8-L      & c $\rightarrow$ b & 49.9 & 37.5 & 68.5 & 22.2 & 23.9 &  -   & 21.9 & 32.3 & 32.0 \\
    CLDA-YOLOv8-L & c $\rightarrow$ b & 61.1 & 52.0 & 75.3 & 39.4 & 40.5 &  -   & 40.5 & 40.2 & 43.7 \\
    \bottomrule
    \end{tabular*}
    \caption{Our proposed framework is applied to YOLOv8 and yields substantial gains in both Cityscapes $\rightarrow $ Foggy Cityscapes and Cityscapes $\rightarrow $ BDD100k experimental settings, the average precision (AP, \%) of the rest categories is presented.}
    \label{tab:v8}
\end{table*}
\noindent\textbf{Quantitative Ablation}. We have conducted sufficient ablation experiments using the model with $640 \times 640$ resolution to prove the effectiveness of the proposed individual components. First, as can be seen from the table, the domain adaptive framework we constructed for the YOLO detector is effective and can greatly improve the performance of the detector migration. 
In addition, the domain alignment based on contrastive learning can significantly improve the accuracy of the student-teacher system, which proves that the alignment we adopt is reasonable.
\begin{table}[htbp]
    \setlength{\abovecaptionskip}{+0.1cm}
    \setlength{\belowcaptionskip}{-0.3cm}
    \setlength{\tabcolsep}{4pt}
    \begin{tabular*}{0.45\textwidth}{ccccccc}
       \toprule
       YOLOv5-L$^{\dagger}$  & ST         & UC         & DP         & GRL        & CA          & mAP@.5 \\
       \midrule
       \checkmark &            &            &            &            &             &  32.5(+0.0)  \\
       \checkmark & \checkmark &            &            &            &             &  44.3(+11.8)  \\
       \checkmark & \checkmark & \checkmark &            &            &             &  46.9(+14.4)  \\
       \checkmark & \checkmark & \checkmark & \checkmark &            &             &  47.6(+15.1)  \\
       \midrule
       \checkmark & \checkmark & \checkmark & \checkmark & \checkmark &             &  48.8(+16.3)  \\
       \checkmark & \checkmark & \checkmark & \checkmark &            &  \checkmark &  49.9(+17.4)  \\
       \checkmark & \checkmark & \checkmark & \checkmark & \checkmark &  \checkmark &  51.6(+19.1)  \\
       \bottomrule
    \end{tabular*}
    \caption{Quantitative ablation study. We report mAP@.5 of each experimental setting and its gains are shown in parentheses. $\dagger$ indicates the use of anchor-free head. ST, UC, DP, and CA denote the teacher-student architecture, uncertainty learning, dynamic data augmentation, and contrastive learning-based alignment proposed above, respectively.}
    \label{tab:4}
\end{table}

\noindent\textbf{Domain Adversarial vs. Contrastive Learning}.
From the values in Table ~ref{tab:4}, the accuracy gain obtained by adding the domain confrontation used for feature alignment is slightly less than that obtained by our proposed CA method used for instance alignment.
However, in multiple training sessions, we found that the model with the added domain adversarial loss was able to bring the teacher-student model into stable learning faster, suggesting that the domain discriminator obfuscates the input features to a certain degree and is able to assist the contrastive learning for efficient domain alignment.
The final result also shows that GRL and CA can work together for better performance (with mAP of 51.6).

\noindent\textbf{Loss function Compared with \cite{khosla2020supervised}}. 
For a fair comparison, we only replace the contrastive loss of CLDA-YOLO with SupCon~\cite{khosla2020supervised}. As previously analyzed, the softmax-based contrastive loss does not work well under domain adaptive object detection, giving only a small gain (+1.1\% mAP). More analysis can be found in the Appendix.
\begin{table}[h]
    \setlength{\abovecaptionskip}{+0.1cm}
    \setlength{\belowcaptionskip}{-0.5cm}
    \centering
    \begin{tabular*}{0.27\textwidth}{c|c}
       \toprule
       Methods     & mAP@.5 \\
       \midrule
       SupContrast~\cite{khosla2020supervised}      & 49.9  \\
       $L_{CA}$ (ours)                              & 51.6  \\
       \bottomrule
    \end{tabular*}
    \caption{Softmax-based supervised contrastive learning loss compared to our proposed sigmoid-based contrastive learning loss.}
    \label{tab:5}
\end{table}

\noindent\textbf{Result on YOLOv8~\cite{yolov8_ultralytics}}.
In view of the fact that our proposed framework uses anchor-free detector, it can be easily combined with the state-of-the-art YOLO. 
In order to verify the generalizability of the proposed algorithm, we performed two sets of experimental validation in YOLOv8. From the table~\ref{tab:v8}, it can be seen that even for the state-of-the-art YOLO detector, our proposed framework can well improve its performance under domain shift.


\noindent\textbf{Qualitative visualization analysis}. 
Fig.~\ref{fig:ap} shows a visualization of CLDA-YOLO (below) compared to the model trained with the source domain only (above), where we have chosen three sets of images from left to right: a regular street scene, a rainy scene, and a dense fog scene. It can be seen that in the leftmost, even with less domain shift, CLDA-YOLO still improves the detection performance of the target domain and provides more accurate detection boxes. In the middle scene, the detection model of the source domain has a large number of errors occur, while CLDA-YOLO basically correctly recognizes the cars appearing in the scene. In the rightmost foggy scene, the source-domain detector has failed to recognize objects on the left side due to occlusion factors, while our proposed algorithm still performs well. More experimental results will be shown in the Appendix. 
We further validate the effectiveness of CA by visualizing the feature distribution of the detection boxes on each of the three detection heads using t-SNE~\cite{van2008visualizing}. As shown in Fig.~\ref{fig:tsne}, the CA strategy makes the object features in two domains forcibly aligned, which mitigates the phenomenon of domain shift. Since CA performs instance-level contrastive learning, dissimilar category distances are pushed farther apart (e.g., car and persion), indicating the applicability of CA for the object detection task.

\section{Conclusion}
In this work, we first construct a domain adaptive architecture for a single-stage YOLO detector and propose to utilize uncertainty learning, dynamic data augmentation, etc. to improve the performance of the algorithm. To achieve domain alignment across multiple sections, we propose Contrastive Align(CA) method that obtains a global representation by maintaining a dynamic queue and performs contrastive learning using Sigmoid-based loss. Our proposed algorithm achieves state-of-the-art performance in multiple dataset settings. 
Since the alignment approach we employ can step outside the binary domain adversarial paradigm, I hope our work can be extended to Multi-target Domain Adaptive tasks to provide more efficient and robust object detection algorithms for the real world in the future.

{
    \small
    \bibliographystyle{ieeenat_fullname}
    \bibliography{main.bib}
}


\end{document}